\documentclass[10pt,a4paper]{article}

\usepackage{cogsci}

\usepackage{hyperref}
\usepackage{booktabs}
\usepackage[table,xcdraw]{xcolor}
\usepackage{makecell}
\usepackage{multirow}
\usepackage{graphicx}
\usepackage{pslatex}
\usepackage{apacite}
\usepackage{float} 
\usepackage{amsmath}

\title{Decoding EEG Signals to Explore Next-Word Predictability in the Human Brain}
 
\author{
\begin{tabular}{ccc}
\textbf{Boi Mai Quach} & \textbf{Binh T. Nguyen} & \textbf{Cathal Gurrin} \\
mai.quach3@mail.dcu.ie & ngtbinh@hcmus.edu.vn & cathal.gurrin@dcu.ie \\
School of Computing, ML-Labs, & Department of Computer Science, & School of Computing, ADAPT Centre,\\
Dublin City University, Ireland & VNUHCM – University of Science & Dublin City University, Ireland \\
\end{tabular}
\\
\\
\hspace{-0.75cm} 
\begin{tabular}{ccc}
 & \textbf{Graham Healy} & \\
 & graham.healy@dcu.ie &  \\
 & School of Computing, ADAPT Centre, &\\
 & Dublin City University, Ireland &
\end{tabular}
}

\begin{document}

\maketitle

\begin{abstract}
Humans invented reading and have passed down this complex skill across generations through language. This study provides empirical evidence of the neural mechanisms underlying bottom-up (related to high-order linguistic structure) and top-down (related to next-word predictability) processes, which interact to guide comprehension during reading. While previous studies have focused on either the N400 effects of predictability or lexical categories, research on how predictability influences N400 responses across different lexical categories is limited, mainly due to constraints in publicly available datasets. Here, we examine how predictability influences brain responses, recorded at millisecond resolution using electroencephalography (EEG), with a focus on the N400 time window (300-500 ms post-stimulus) across different lexical and grammatical categories. Our results indicate that significant differences in N400 responses between high and low cloze probability levels were more pronounced for content words than function words. Among the two primary content categories, verbs exhibited greater N400 differences than nouns, while nouns carried more distinct information about their predictability than verbs. Moreover, we demonstrate that the decoding technique is more effective than the event-related potential (ERP) traditional analysis in capturing more detailed and distinct representations of cognitive processes over time.

\end{abstract}

\section{Introduction}
Reading comprehension involves two interconnected processes: bottom-up and top-down processing \cite{perfetti2001lexical}. While bottom-up processing is often considered an automatic outcome of accurate word recognition \cite{macdonald1994lexical,ngabut2015reading}, top-down processing uses cues to disambiguate and predict upcoming words \cite{federmeier2007thinking,kuperberg2016we}. For example, we can infer that the statement ``I drink my coffee with cream and ...'' is likely to end with ``sugar.''. Readers access linguistic structures (e.g., lexical groups, grammatical categories) via a bottom-up process to predict that the next word is probably a noun. Simultaneously, they rely on top-down cues to anticipate the upcoming word with its correct meaning before it is fully revealed.

The N400 ERP component is a negative-going deflection peaking around 400 ms post-stimulus word onset, well known for its sensitivity to semantic complexity \cite{kutas2011thirty}. Cloze probability \cite{taylor1953cloze}, which estimates how likely a particular word is in a given sentence and also reflects the top-down prediction process, has long been considered the primary predictor of N400 amplitude. The graded attenuation of the N400 as a function of cloze probability is one of the most widely replicated phenomena in reading comprehension using EEG, supporting the conclusion that words with higher cloze value elicit a smaller (less negative) N400 response compared to words with lower cloze value \cite{federmeier1999rose, block2010cloze, michaelov2022so}.

Basically, words are grouped into two primary types: content and function groups. Content words, including nouns, verbs, adjectives, and most adverbs, convey specific semantic information about objects and events. Function words, on the other hand, comprise pronouns, determiners, auxiliaries, and adpositions and serve as structural elements in phrase construction. Evidence from ERP recordings during reading processing supports a lexical-based analysis guided by a bottom-up process. Results from pioneering studies indicate that content words elicit a larger N400 amplitude compared to function words \cite{neville1992fractionating}. This finding was later supported by observations showing that both word classes could elicit N400 responses, but function words, especially high-frequency ones, show significantly smaller amplitudes \cite{munte2001differences,he2022neural}. Grammatical categories, particularly nouns and verbs, also influence N400 amplitude, with larger amplitudes observed for verbs compared to nouns \cite{federmeier2000brain, lee2006mind}.

There is a lack of evidence in lexical-based analysis using next-word predictability due to the limited availability of publicly accessible EEG datasets for semantic-level information \cite{deniz2019representation}. To gain deeper theoretical insights, this study aims to test two hypotheses: The first hypothesis states that content words elicit a larger N400 difference effect between high and low cloze levels compared to function words. The second hypothesis suggests that the N400 predictability difference between high and low cloze is more pronounced for verbs than nouns.

Moreover, we not only explore the nature of top-down processing and its relationship to bottom-up information processing during reading comprehension, but we also provide evidence that multivariate decoding outperforms traditional ERP analysis in capturing distinct neural information across different conditions. By leveraging univariate and decoding techniques to investigate our hypotheses, we further demonstrate the advantages of multivariate decoding in this context.

\section{Methodology}

\subsection{EEG Recording and Data Preparation}

To examine the N400 response to the predictability across different lexical classes, we utilised the DERCo dataset \cite{quach2024derco}, including EEG data recorded from 22 native English speakers while they were reading The Grimm Brothers’ Fairy Tales. Two participants ``QPF42'' and ``USQ95'' were excluded from data analysis due to excessive eye movements in their data. Additionally, word-by-word cloze values were collected using a cloze procedure via Mechanical Turk crowdsourcing platform. High-density EEG data was recorded using electrode 32-channels placed according to the international 10–20 system \cite{klem1999ten}.

Parts of speech were identified using the Python library IPA \footnote{https://github.com/Riccorl/ipa}, which allows us to extract and categories the grammatical function of each word within a sentence. For example, in the sentence ``She quickly reviewed some detailed reports with her team,'' the parts of speech are categorised as follows: She (pronoun), quickly (adverb), reviewed (verb), some (determiner), detailed (adjective), reports (noun), with (adposition), her (pronoun), team (noun). In terms of cloze probability, we utilised the same two distinct groups (high versus low) as in the DERCo dataset. This decision was based on their validation, which indicated significant differences in N400s across most electrodes.

After preprocessing, the number of trials for high and low cloze probability levels in terms of lexical classes varied between participants. To standardise the trial numbers for analysis, we identified the participant with the fewest trials at each cloze probability level and set these as a baseline. We then randomly selected an equivalent number of trials from other participants,
ensuring their cloze value distributions closely matched the baseline. This random selection process was guided by the Kolmogorov-Smirnov (KS) test, iterating up to 5000 times to find the subset with the lowest KS statistic, indicating the best distribution match. As mentioned, our analysis focused on binary classifications such as lexical groups (content versus function words) and content word categories (nouns versus verbs). The number of trials used as a baseline for each group per condition is detailed in Table \ref{table:mass_univariate_analysis}.

\subsection{Univariate Analysis}
\label{sec:univariate}

Traditional univariate analysis involves calculating a difference wave between two or more conditions for a given ERP component. The most common statistical approach to dealing with this issue uses data clustering combined with permutation testing \cite{maris2007nonparametric}. By examining data across all electrode sites, these analyses ensure comprehensive spatial coverage, increase the statistical power, control for multiple comparisons, and allow for the identification of a cluster of electrodes where the component of interest is largest \cite{maris2007nonparametric,groppe2011mass,luck2017get}. In the standard univariate approach, the difference in ERP amplitude between the two conditions is quantified at multiple time points. For the mass univariate analyses, we have a significantly larger set of (channel, time) pairs, also referred to as samples, wherein we aim to examine the N400 effect. 

In our analysis, each word-level EEG epoch will have three dimensions: 32 electrodes x 20 participants x 600 time points (ranging from $-100$ to $500$ ms after word onset), requiring multiple comparisons. In comparison to single-sensor analyses, the multiple comparisons problem is significantly more pronounced in this context: with 32 channels and 600 temporal points, we generate 19,200 $t$-values. The cluster-based permutation test (10,000 times) used a point-wise independent samples t-test to identify clusters with data points below the alpha level ($p < 0.05$). For multi-channel analyses, the approach is similar to single-channel analyses but differs in clustering: rather than clustering based solely on temporal adjacency, and we now cluster the selected (channel, time) samples based on both spatial and temporal adjacency. 

A major advantage of this analysis is its capability to identify the optimal group of electrode sites within a specific time window. Once clusters of (channel, time) samples are identified, those with $p$-values exceeding the critical two-sided alpha-level (0.05) are considered insignificant. The false discovery rate (FDR) is calculated and used to correct for multiple comparisons \cite{riffenburgh2020statistics}. Each significant cluster represents a contiguous block of activity across time points and potentially across electrode channels. By averaging the difference waves for all participants within an optimal cluster of electrode sites, we can approximate the electrode locations that best account for the ERP effect.  Note that if the approach does not yield any optimal electrode sites using FDR correction, we will select optimal channels without applying the correction. Table~\ref{table:mass_univariate_analysis} shows the results of the optimal clusters of electrode sites in the N400 time window.

\begin{table}[!ht]
\setlength{\tabcolsep}{4pt} 
\renewcommand{\arraystretch}{1.3} 

\begin{tabular}{|l|c|c|p{4cm}|} 
    \hline
    \multirow{2}{*}{\textbf{Group}} & \multicolumn{2}{c|}{\textbf{Baseline Trials}} & \multirow{2}{*}{\textbf{Optimal Electrode Cluster}} \\\cline{2-3}
    & \textbf{High} & \textbf{Low} & \\\hline
    \textbf{Content} & 193 & 365 & P3, P7, CP1, CP2, Pz, P4, Fp2, F7, F3, Fz, F4 \\\hline
    \textbf{Function} & 318 & 155 & FT10, FC1, FC2, C3, Cz, C4, Fp2, F7, F3, Fz, F4 \\\hline
    \textbf{Noun} & 125 & 93 & P3, P7, CP1, CP2, Pz, P4, F7, F3, Fz, F4 \\\hline
    \textbf{Verb} & 40 & 165 & T8, CP6, FT9, P3, P7, CP1, CP2, Pz, P4, F7, F3, Fz, F4 \\\hline
\end{tabular}

\caption{Baseline trial counts and optimal electrode clusters in the N400 time window for each condition. \textbf{High} and \textbf{Low} refer to high and low cloze conditions.}
\label{table:mass_univariate_analysis}
\end{table}

\subsection{Decoding Analysis}

Compared to other Machine Learning (ML) models, such as linear discriminant analysis and random forests, support vector machine (SVM) has demonstrated superior performance in examining N400 effects of prediction and semantic relatedness \cite{trammel2023decoding}. SVM is particularly effective in decoding EEG/ERP data due to its ease of implementation, strong performance with small training sets, and especially its ability to handle non-linear relationships in high-dimensional spaces \cite{carrasco2024using} by using kernel functions. 

\begin{figure*}[!ht]
    \centering
    \includegraphics[width=1\textwidth, height = 0.36\textwidth]{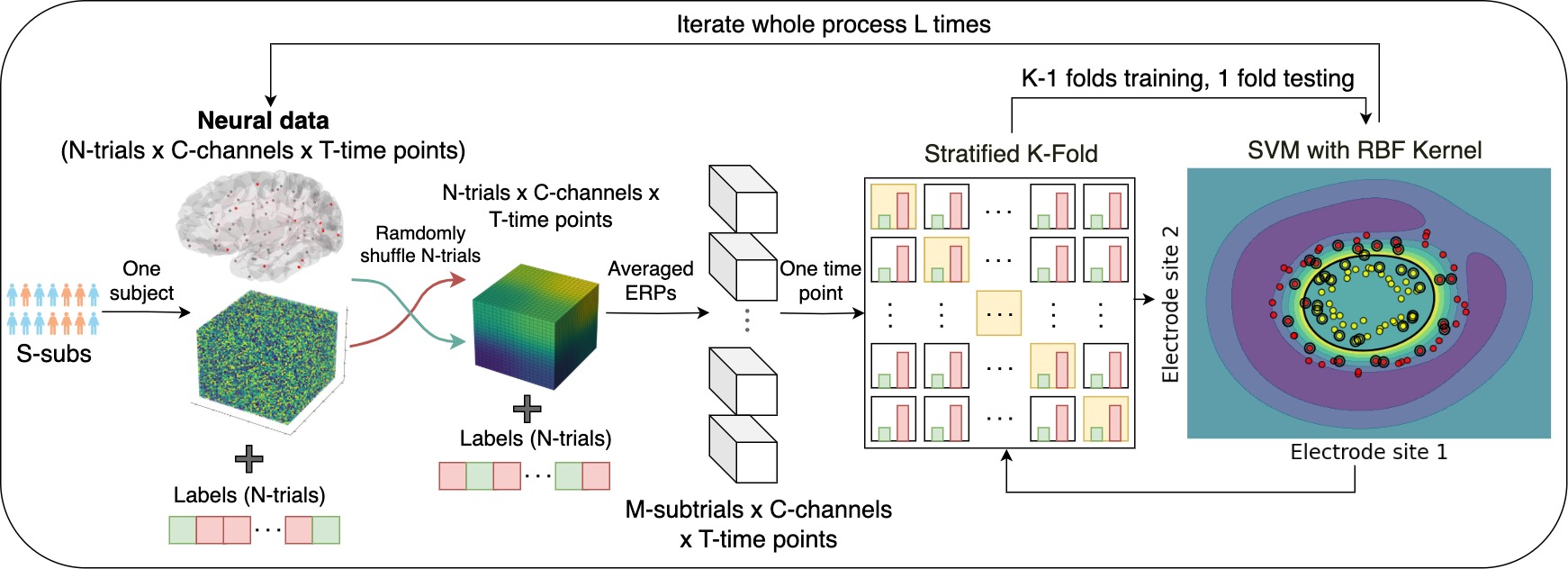}
    \caption{Flowchart of the EEG decoding process using an SVM with an RBF kernel. The process begins with EEG data from each participant, organized into 3D arrays by N trials, C electrodes, and T time points. Trials are randomly shuffled to reduce bias, and the SNR is increased by averaging certain EEG trials per set to create data with M sub-trials, C electrodes, and T time points. Stratified K-fold cross-validation is applied to balance training and testing data, followed by SVM training and testing on each fold. The process is repeated L times per participant, with a total of S participants.}
    \label{fig:flowchart_decoding}
\end{figure*}

The flowchart in Figure~\ref{fig:flowchart_decoding} illustrates EEG data decoding using an SVM with an RBF kernel. The process begins with the collection of neural data from multiple subjects, where each participant's data is organised into a 3-dimensional array corresponding to trials, electrode sites, and time points. To reduce the potential bias, the trials are randomly shuffled, mitigating order effects and ensuring better generalisation. Single-trial EEG epochs are often too noisy to effectively decode subtle stimulus classes. An increased signal-to-noise ratio (SNR) can be achieved by randomly dividing the data into M sets of approximately a certain number of trials each, creating an averaged ERP for each set \cite{isik2014dynamics,carrasco2024using}. Many previous studies \cite{isik2014dynamics,grootswagers2017decoding,carrasco2024using} have used multiple sets of 10–20 trials and achieved the highest or near-highest performance in classifier accuracy. Therefore, an average of 15 trials was utilised in this study. For example, 80 high- and 80 low-cloze for the content group could be averaged into five sets of 16 trials per condition. This approach also captures within-subject variability and enhances decoding reliability.

The averaged ERPs are then subjected to stratified K-fold cross-validation \cite{bishop2006pattern}, a technique that addresses the unbalanced distribution of classes and minimizes overfitting risks \cite{bae2018dissociable, carrasco2024using}, ensuring each class receives fair representation in both training and validation phases. In our study, the SVM-based decoding was repeated five times. For each round, an SVM classifier with an RBF kernel was trained on four folds and tested on the remaining fold. As in the main decoding procedure, this procedure was applied to each time point independently. Decoding accuracy is defined as the proportion of test cases that are correctly classified. To increase the resolution of the decoding accuracy, the entire process is iterated 100 times (L) for each participant. For each iteration, we re-randomised the assignment of trials to averages and trained new SVMs.

If the neural patterns encode distinct lexical class information, decoding accuracy should be greater than the 0.5 chance level for binary classifications. To compare decoding accuracy to the chance at each time point while controlling for multiple comparisons, we used a cluster-based permutation technique similar to that in the univariate analysis. We used a one-sample $t$ test to compare the mean accuracy across participants to chance. However, one-tailed tests were used for the SVM decoding accuracies instead of two-tailed tests in the univariate technique because the SVM classifier should not produce meaningful below-chance decoding. 

For a given score (i.e., a voltage in a univariate difference wave or a decoding accuracy), we selected a random sample of n of the N participants (sampling from the set of N participants with replacement) and computed the effect size for this random sample. We then conducted 5,000 iterations of the testing procedure, making it possible to construct the null distribution of the maximum cluster-level $t_{max}$ with a cluster $p$-value. 

\subsection{Comparison of Performance Between Methods}
Both SVM-based decoding and univariate analysis showed significant N400 amplitude differences between lexical groups but could not quantify effect magnitude or directly compare ERP amplitude variations. A common solution is to use effect size or statistical power via Cohen's $d_z$ \cite{cohen2013statistical}. Effect size measures the magnitude of the difference between outcomes. This approach allows researchers to present the magnitude of reported effects in a standardised metric, which can be understood regardless of the scale used to measure the dependent variable \cite{lakens2013calculating}.

This metric quantifies the ability of each approach to produce statistically significant results while accounting for variation in wave difference (in univariate analysis) and decoding accuracy (in SVM-based decoding analysis) across participants as well as the mean. A larger effect size reduces the Type II error rate, thereby increasing the proportion of significant effects \cite{ioannidis2005most}. Thus, the approach that generated a larger Cohen's $d_z$ exhibited greater statistical power for detecting differences between experimental conditions in a within-subjects analysis. The Cohen's $d_z$ is calculated as:
\begin{equation*}
    d_z = \frac{\bar{X} - \text{chance}}{\sigma_{X}}   
\end{equation*}

In our analysis, regarding the SVM-based decoding approach, $X$ represents the mean decoding accuracy across participants, with a chance level of 0.5, and $\sigma_{X}$ denotes the standard deviation of these accuracy values. For the univariate analysis, $X$ represents the mean wave difference between two conditions across participants, with a chance level of 0, and $\sigma_{X}$ denotes the standard deviation of these difference values. Bootstrapping (10,000 iterations) is then used to estimate the standard error of the effect size. 

\section{Results}

\subsection{Content vs. Function Words}
\subsubsection{Traditional Univariate Analysis} 

\begin{figure*}[!ht]
    \centering
    \includegraphics[width=0.92\textwidth, height = 0.25\textwidth]{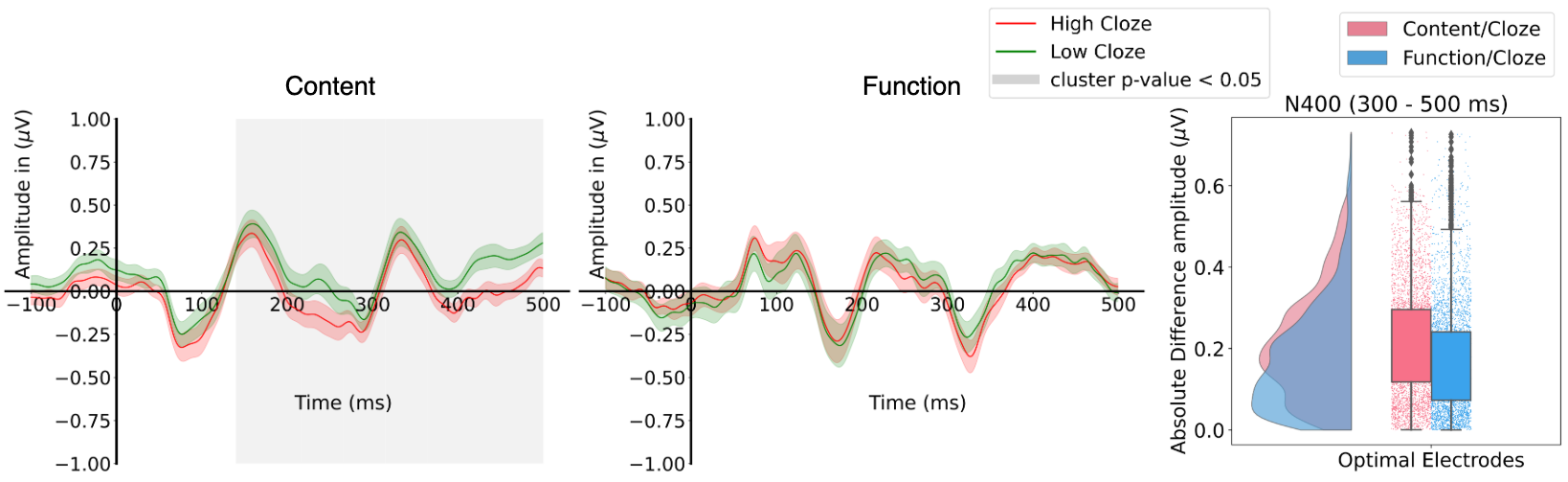}
    \caption{
    ERPs plotted for content words (leftmost) and function words (middle) with high cloze (red) and low cloze (green). Gray shading indicates significant time points (p-values $< 0.05$, two-tailed). The raincloud plots (rightmost) show the absolute N400 amplitude difference between content/cloze (pink) and function/cloze (blue) conditions across optimal electrodes.
    }
    \label{fig:univariate_lexical_analysis}
\end{figure*}

Figure \ref{fig:univariate_lexical_analysis} summarizes the performance of the N400 differences (measured in $\mu V$) between high and low cloze in both lexical classes. Results from the temporal permutation cluster test show that the N400 differences are statistically significant for content words from 150 to 500 ms after word onset ($p < .05$). In contrast, no significant difference in N400 predictability effects was observed for function words relative to word onset after correction for multiple comparisons. Indeed, function words have privileged access to prediction, especially when they are very frequent \cite{bell2009predictability}, making their N400 responses less sensitive to cloze probability differences. Conversely, content words, being generally less frequent and more concrete in meaning \cite{bell2009predictability}, show greater sensitivity to such differences.

\subsubsection{Decoding Analysis} 

\begin{figure*}[!ht]
    \centering
    \includegraphics[width=0.92\textwidth, height = 0.26\textwidth]{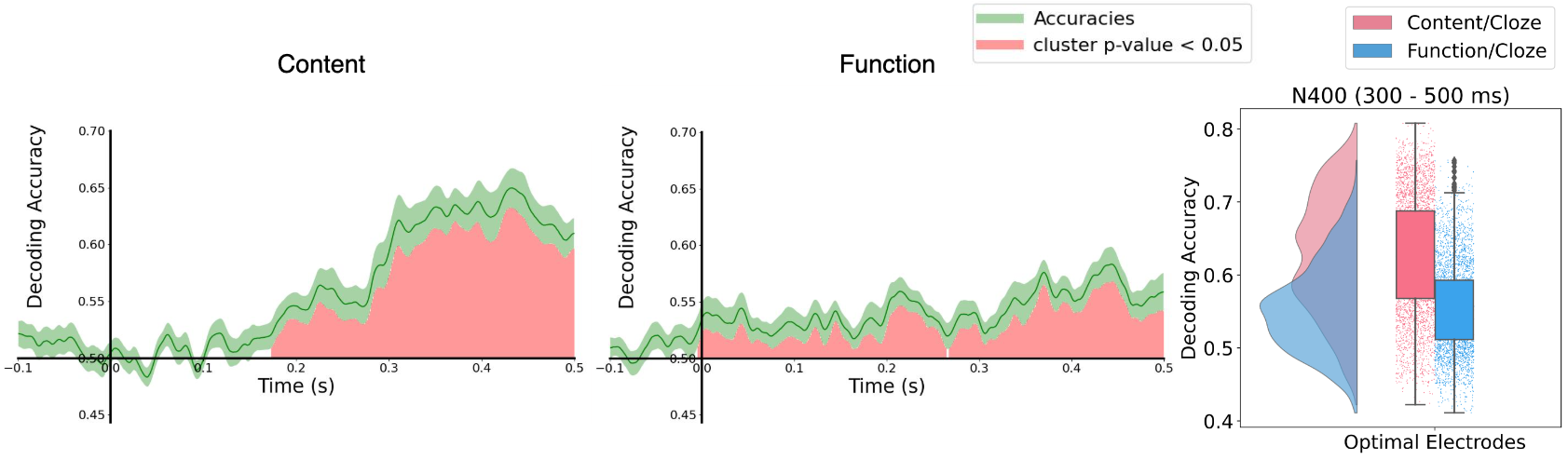}
    \caption{Decoding accuracy plotted for content words (leftmost), function words (middle) with high cloze (red) and low cloze (green). Red shading indicates significant time points (p-values $< 0.05$, one-tailed). The raincloud plots (rightmost) show the decoding accuracy for content/cloze condition (pink) and function/cloze condition (blue) across optimal electrodes.}
    \label{fig:decoding_lexical_analysis}
\end{figure*}

Figure \ref{fig:decoding_lexical_analysis} shows that word class was decodable from the neural responses during the entire epoch ($p < .05$).
Interestingly, the SVM decoding demonstrated strong performance, with accuracy consistently exceeding chance levels (0.5), except for a very few time points in the early time course where it slightly dropped below chance, likely accounted for by random noise rather than systematic misclassification.

Decoding accuracy for content words increased notably over time, peaking at 300-500 ms after word offset, with a maximum accuracy of around 0.65. Although function words were also encoded in neural activity, their decoding accuracy remained around 0.55. This difference between content and function words was possibly accounted for by cloze predictability of the upcoming word, in which neural activity increasingly differentiates between high and low cloze content words, while function words, which are more syntactically oriented and carry less semantic weight, generate less distinctive neural responses based on cloze probability.

\subsubsection{Comparisons} 

Both approaches effectively detect neural differences between high and low cloze probability words but offer different insights. The univariate analysis highlights the amplitude difference of the N400 component but struggles to differentiate between content and function words. In contrast, SVM-based decoding provides a more dynamic and precise measure of how the brain processes word predictability over time. The raincloud plots (rightmost) showing the absolute N400 amplitude (Figure \ref{fig:univariate_lexical_analysis}) and the decoding accuracy (Figure \ref{fig:decoding_lexical_analysis}) indicate that the distinction is easier to observe through decoding results than through traditional ERP analysis.

\subsection{Nouns vs. Verbs}
\subsubsection{Traditional Univariate Analysis} 
\begin{figure*}[!ht]
    \centering
    \includegraphics[width=0.92\textwidth, height = 0.26\textwidth]{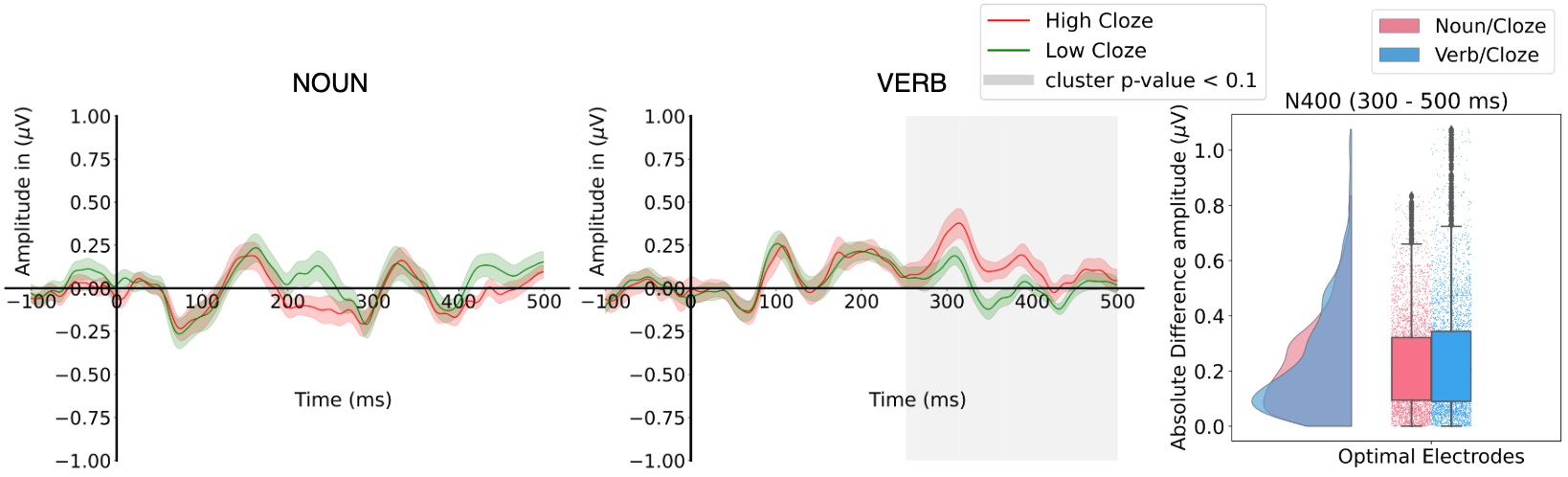}
    \caption{
    ERPs plotted for nouns (leftmost) and verbs (middle) with high cloze (red) and low cloze (green). Gray shading indicates significant time points (p-values $< 0.1$, two-tailed). The raincloud plots (rightmost) show the absolute N400 amplitude difference between nouns/cloze (pink) and verbs/cloze (blue) conditions across optimal electrodes.}
    \label{fig:content_univariate_analysis}
\end{figure*}

Figure \ref{fig:content_univariate_analysis} examines how nouns and verbs are processed under varying levels of predictability (high vs. low cloze probability) within their optimal electrode clusters. Using the permutation cluster approach at a threshold of $p < .05$, we did not find significant clusters for either nouns or verbs. However, at a more liberal threshold of $p < .1$, verbs showed a significant difference in the N400 time window, whereas nouns still showed no neural differences between predictability conditions. This finding suggests that verbs may be more sensitive to contextual predictability than nouns, likely due to the greater complexity and flexibility of verb semantics in varying contexts \cite{matzig2009noun,earles2017verbs}. To obtain more robust results, we plan to collect additional cloze data in the next version of the DERCo dataset, increasing statistical power for ERP analyses across grammatical categories.

\subsubsection{Decoding Analysis} 

\begin{figure*}[!ht]
    \centering
    \includegraphics[width=0.92\textwidth, height = 0.26\textwidth]{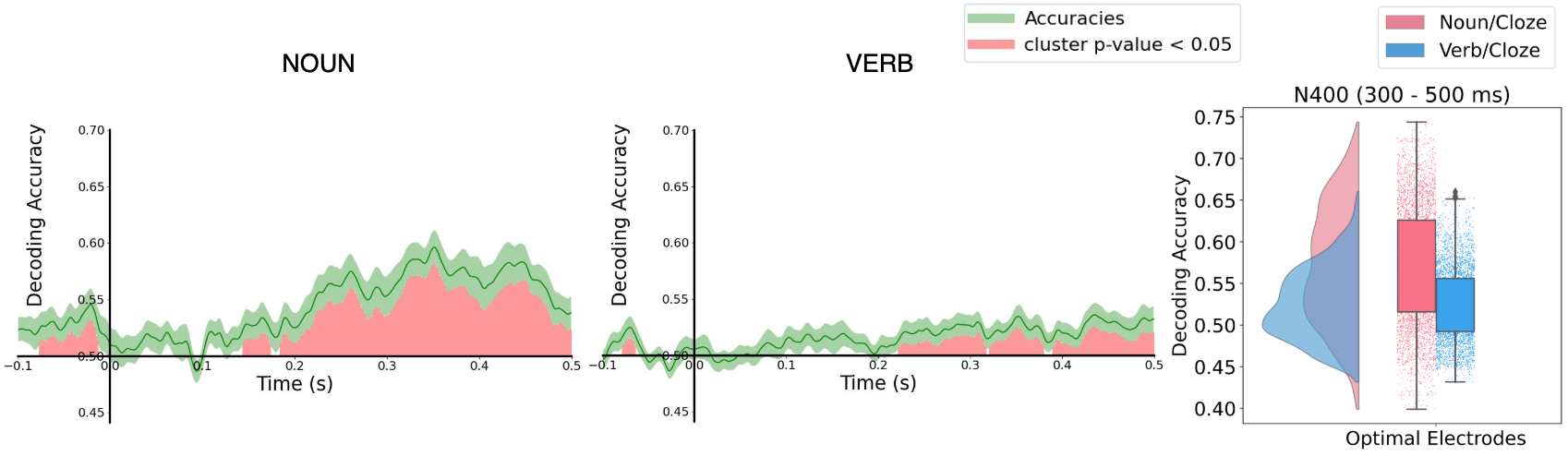}
    \caption{
    Decoding accuracy plotted for nouns (leftmost), verbs (middle) with high cloze (red) and low cloze (green). Red shading indicates significant time points (p-values $< 0.05$, one-tailed). The raincloud plots (rightmost) show the decoding accuracy for nouns/cloze condition (pink) and verbs/cloze condition (blue) across optimal electrodes.
    }
    \label{fig:decoding_content_class}
\end{figure*}

The results in Figure \ref{fig:decoding_content_class} indicate that nouns acquired a higher decoding performance than verb, suggesting that neural signals for nouns carry more distinct information regarding their predictability across high and low cloze word groups, particularly during the N400 time window (accuracy $> 10\%$). This richer representation might be due to the generally more stable and predictable nature of nouns in language, making them easier to decode at the neural level \cite{vigliocco2011nouns}. Verbs, however, show lower overall decoding accuracy, reflecting more complex and distributed processing required for integrating verbs into the sentence structure \cite{bird2000verb,earles2017verbs}.

\subsubsection{Comparisons} 

Nouns, despite showing a less pronounced N400 amplitude differences between high and low cloze in the univariate analysis, carry more distinct and accessible neural information than verbs, as outlined in decoding accuracy. This discrepancy arises because traditional ERP analyses quantify neural response differences between conditions while assuming a consistent scalp distribution across participants. In contrast, decoding analyses assess the amount of information about a condition present in the recorded signal for each participant individually without assuming uniform brain activity across participants. This approach allows decoding to uncover more aspects that might be missed by applying traditional ERP methods.

\subsection{Results of Size Effects}
\begin{figure}[!ht]
    \centering
    \includegraphics[width=0.495\textwidth, height = 0.33\textwidth]{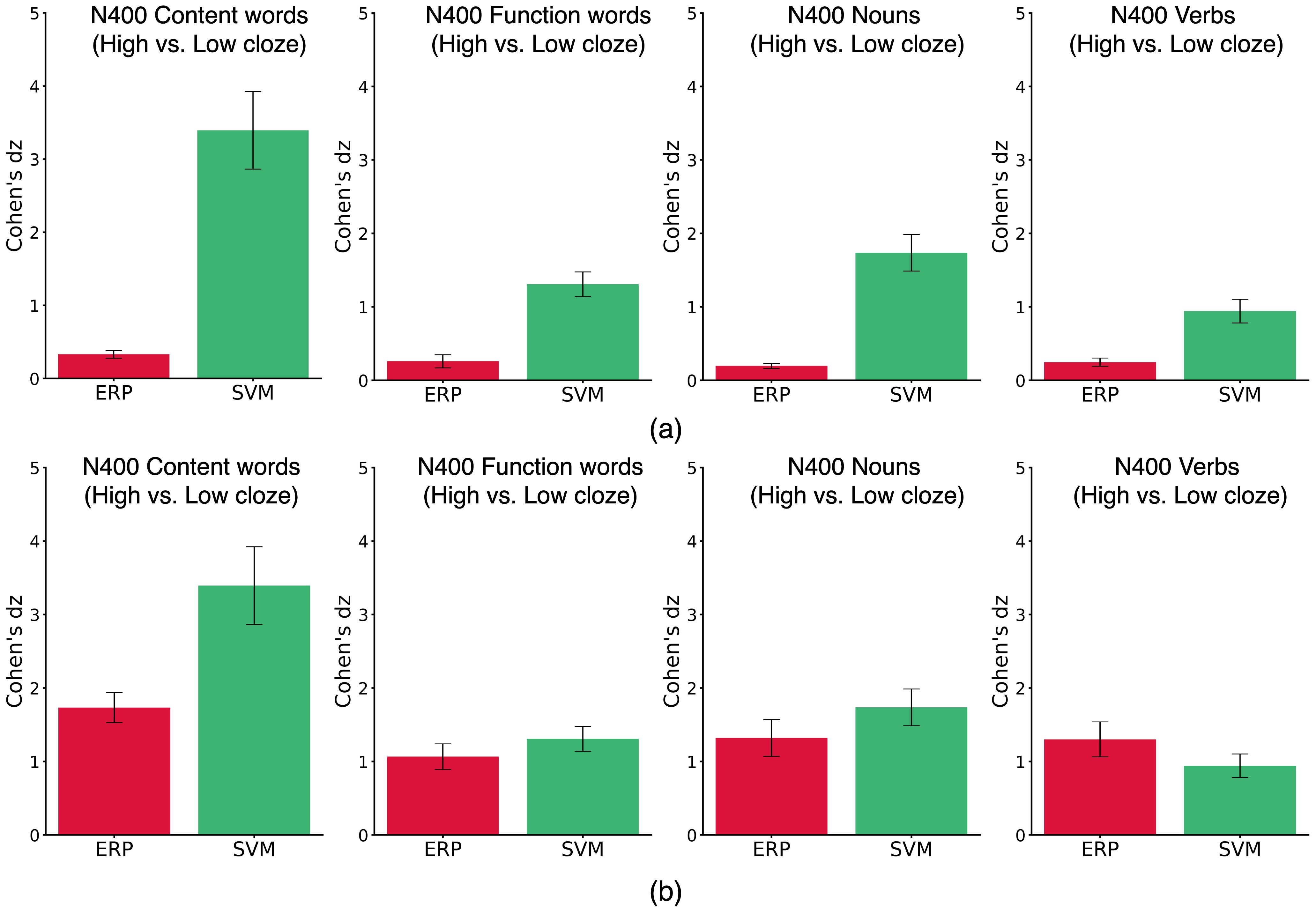}
    \caption{Effect size (Cohen's $d_z$) plotted on the same scale for each analysis in the N400 time window, comparing the traditional ERP analysis and the decoding technique, with (a) all electrodes and (b) optimal electrodes. The error bars show $\pm1$ standard error.}
    \label{fig:cohen_dz_analysis}
\end{figure}

Figure \ref{fig:cohen_dz_analysis} provides a comparative analysis of N400 effect sizes between SVM-based decoding and traditional ERP analysis across four cases: high versus low cloze probabilities for content words, function words, nouns, and verbs. In the univariate approach, the difference in ERP amplitude between the two conditions was quantified at a single electrode site. To increase effect size in conventional ERP analysis, we averaged the difference waves across 32 electrode sites.

In Figure \ref{fig:cohen_dz_analysis} (a), SVM-based decoding consistently outperformed conventional ERP analysis in effect sizes across all cases. To increase the effect size of traditional ERP analysis, we used the optimal group of electrode sites that best accounts for the N400 effect (see Table~\ref{table:mass_univariate_analysis}). As shown in Figure~\ref{fig:cohen_dz_analysis} (b), the decoding technique still demonstrated superior sensitivity and effect sizes despite a significant improvement in ERP effect sizes. These results highlight the potential advantages of employing multivariate decoding techniques in neurocognitive language research. 

\section{Discussion and Conclusion}

By leveraging both traditional ERP analysis and decoding techniques, we have provided compelling evidence for the neural mechanisms of next-word prediction in reading through N400 analysis. Our results show that content words, carrying more semantic information, evoke stronger N400 effects than function words, which primarily aid in syntactic structuring, supporting prior research on the greater engagement of semantic processing in content words. Moreover, the decoding technique offers valuable theoretical insights, revealing that although verbs exhibit greater N400 differences than nouns, nouns carry more distinct predictability information. The comparison between two techniques highlights the advantages of multivariate techniques in capturing neural complexity. While univariate approaches identify broad patterns, they often miss subtle differences in brain activity.

While this study focused on the N400 component, our ongoing research explores additional ERP components, such as the P600, to gain further insights into syntactic processing \cite{gouvea2010linguistic} and provide a more holistic understanding of the neural mechanisms involved in reading comprehension. Likewise, future research should consider using additional decoding techniques to compare their results and potentially gain a more comprehensive understanding of the neural data \cite{trammel2023decoding,carrasco2024using}.

\section{Acknowledgements}
This publication has emanated from research conducted with the financial support of Science Foundation Ireland under Grant number 18/CRT/6183 and 13/RC/2106\_P2. We would like to thank the anonymous reviewers for their helpful remarks.

\section{Code Availability}
All the scripts for analysis can be found at \url{https://github.com/Tayerquach/brain_decoding_model}.

\bibliographystyle{apacite}

\setlength{\bibleftmargin}{.125in}
\setlength{\bibindent}{-\bibleftmargin}

\bibliography{CogSci}

@article{macdonald1994lexical,
  title={The lexical nature of syntactic ambiguity resolution.},
  author={MacDonald, Maryellen C and Pearlmutter, Neal J and Seidenberg, Mark S},
  journal={Psychological review},
  volume={101},
  number={4},
  pages={676},
  year={1994},
  publisher={American Psychological Association}
}

@article{ngabut2015reading,
  title={Reading theories and reading comprehension},
  author={Ngabut, Maria Novary},
  journal={Journal on English as a Foreign Language},
  volume={5},
  number={1},
  pages={25--36},
  year={2015}
}

@article{federmeier2007thinking,
  title={Thinking ahead: The role and roots of prediction in language comprehension},
  author={Federmeier, Kara D},
  journal={Psychophysiology},
  volume={44},
  number={4},
  pages={491--505},
  year={2007},
  publisher={Wiley Online Library}
}

@article{kuperberg2016we,
  title={What do we mean by prediction in language comprehension?},
  author={Kuperberg, Gina R and Jaeger, T Florian},
  journal={Language, cognition and neuroscience},
  volume={31},
  number={1},
  pages={32--59},
  year={2016},
  publisher={Taylor \& Francis}
}

@article{perfetti2001lexical,
  title={The lexical basis of comprehension skill.},
  author={Perfetti, Charles A and Hart, Lesley},
  year={2001},
  publisher={American Psychological Association}
}

@article{michaelov2022so,
  title={So cloze yet so far: N400 amplitude is better predicted by distributional information than human predictability judgements},
  author={Michaelov, James A and Coulson, Seana and Bergen, Benjamin K},
  journal={IEEE Transactions on Cognitive and Developmental Systems},
  volume={15},
  number={3},
  pages={1033--1042},
  year={2022},
  publisher={IEEE}
}

@article{federmeier1999rose,
  title={A rose by any other name: Long-term memory structure and sentence processing},
  author={Federmeier, Kara D and Kutas, Marta},
  journal={Journal of memory and Language},
  volume={41},
  number={4},
  pages={469--495},
  year={1999},
  publisher={Elsevier}
}

@article{taylor1953cloze,
  title={{“Cloze procedure”: A new tool for measuring readability}},
  author={Taylor, Wilson L},
  journal={Journalism quarterly},
  volume={30},
  number={4},
  pages={415--433},
  year={1953},
  publisher={SAGE Publications Sage CA: Los Angeles, CA}
}

@article{deniz2019representation,
  title={{The representation of semantic information across human cerebral cortex during listening versus reading is invariant to stimulus modality}},
  author={Deniz, Fatma and Nunez-Elizalde, Anwar O and Huth, Alexander G and Gallant, Jack L},
  journal={Journal of Neuroscience},
  volume={39},
  number={39},
  pages={7722--7736},
  year={2019},
  publisher={Soc Neuroscience}
}

@article{block2010cloze,
  title={Cloze probability and completion norms for 498 sentences: Behavioral and neural validation using event-related potentials},
  author={Block, Cady K and Baldwin, Carryl L},
  journal={Behavior research methods},
  volume={42},
  number={3},
  pages={665--670},
  year={2010},
  publisher={Springer}
}

@article{bell2009predictability,
  title={Predictability effects on durations of content and function words in conversational English},
  author={Bell, Alan and Brenier, Jason M and Gregory, Michelle and Girand, Cynthia and Jurafsky, Dan},
  journal={Journal of Memory and Language},
  volume={60},
  number={1},
  pages={92--111},
  year={2009},
  publisher={Elsevier}
}

@article{matzig2009noun,
  title={Noun and verb differences in picture naming: Past studies and new evidence},
  author={M{\"a}tzig, Simone and Druks, Judit and Masterson, Jackie and Vigliocco, Gabriella},
  journal={Cortex},
  volume={45},
  number={6},
  pages={738--758},
  year={2009},
  publisher={Elsevier}
}

@article{earles2017verbs,
  title={Why are verbs so hard to remember? Effects of semantic context on memory for verbs and nouns},
  author={Earles, Julie L and Kersten, Alan W},
  journal={Cognitive science},
  volume={41},
  pages={780--807},
  year={2017},
  publisher={Wiley Online Library}
}

@article{neville1992fractionating,
  title={Fractionating language: Different neural subsystems with different sensitive periods},
  author={Neville, Helen J and Mills, Debra L and Lawson, Donald S},
  journal={Cerebral cortex},
  volume={2},
  number={3},
  pages={244--258},
  year={1992},
  publisher={Oxford University Press}
}

@article{munte2001differences,
  title={Differences in brain potentials to open and closed class words: Class and frequency effects},
  author={M{\"u}nte, Thomas F and Wieringa, Bernardina M and Weyerts, Helga and Szentkuti, Andras and Matzke, Mike and Johannes, S{\"o}nke},
  journal={Neuropsychologia},
  volume={39},
  number={1},
  pages={91--102},
  year={2001},
  publisher={Elsevier}
}

@article{he2022neural,
  title={Neural correlates of word representation vectors in natural language processing models: Evidence from representational similarity analysis of event-related brain potentials},
  author={He, Taiqi and Boudewyn, Megan A and Kiat, John E and Sagae, Kenji and Luck, Steven J},
  journal={Psychophysiology},
  volume={59},
  number={3},
  pages={e13976},
  year={2022},
  publisher={Wiley Online Library}
}

@article{federmeier2000brain,
  title={Brain responses to nouns, verbs and class-ambiguous words in context},
  author={Federmeier, Kara D and Segal, Jessica B and Lombrozo, Tania and Kutas, Marta},
  journal={Brain},
  volume={123},
  number={12},
  pages={2552--2566},
  year={2000},
  publisher={Oxford University Press}
}

@article{lee2006mind,
  title={To mind the mind: An event-related potential study of word class and semantic ambiguity},
  author={Lee, Chia-lin and Federmeier, Kara D},
  journal={Brain Research},
  volume={1081},
  number={1},
  pages={191--202},
  year={2006},
  publisher={Elsevier}
}

@article{quach2024derco,
  title={DERCo: A Dataset for Human Behaviour in Reading Comprehension Using EEG},
  author={Quach, Boi Mai and Gurrin, Cathal and Healy, Graham},
  journal={Scientific Data},
  volume={11},
  number={1},
  pages={1104},
  year={2024},
  publisher={Nature Publishing Group UK London}
}

@article{klem1999ten,
  title={{The ten-twenty electrode system of the international federation. The international federation of clinical neurophysiology}},
  author={Klem, George H},
  journal={Electroencephalogr. Clin. Neurophysiol. Suppl.},
  volume={52},
  pages={3--6},
  year={1999}
}

@article{maris2007nonparametric,
  title={Nonparametric statistical testing of EEG-and MEG-data},
  author={Maris, Eric and Oostenveld, Robert},
  journal={Journal of neuroscience methods},
  volume={164},
  number={1},
  pages={177--190},
  year={2007},
  publisher={Elsevier}
}

@article{groppe2011mass,
  title={Mass univariate analysis of event-related brain potentials/fields II: Simulation studies},
  author={Groppe, David M and Urbach, Thomas P and Kutas, Marta},
  journal={Psychophysiology},
  volume={48},
  number={12},
  pages={1726--1737},
  year={2011},
  publisher={Wiley Online Library}
}

@article{luck2017get,
  title={How to get statistically significant effects in any ERP experiment (and why you shouldn't)},
  author={Luck, Steven J and Gaspelin, Nicholas},
  journal={Psychophysiology},
  volume={54},
  number={1},
  pages={146--157},
  year={2017},
  publisher={Wiley Online Library}
}

@book{riffenburgh2020statistics,
  title={Statistics in medicine},
  author={Riffenburgh, Robert H and Gillen, Daniel L},
  year={2020},
  publisher={Academic press}
}

@article{carrasco2024using,
  title={Using multivariate pattern analysis to increase effect sizes for event-related potential analyses},
  author={Carrasco, Carlos Daniel and Bahle, Brett and Simmons, Aaron Matthew and Luck, Steven J},
  journal={Psychophysiology},
  pages={e14570},
  year={2024},
  publisher={Wiley Online Library}
}

@article{trammel2023decoding,
  title={Decoding semantic relatedness and prediction from EEG: A classification method comparison},
  author={Trammel, Timothy and Khodayari, Natalia and Luck, Steven J and Traxler, Matthew J and Swaab, Tamara Y},
  journal={NeuroImage},
  volume={277},
  pages={120268},
  year={2023},
  publisher={Elsevier}
}

@article{isik2014dynamics,
  title={The dynamics of invariant object recognition in the human visual system},
  author={Isik, Leyla and Meyers, Ethan M and Leibo, Joel Z and Poggio, Tomaso},
  journal={Journal of neurophysiology},
  volume={111},
  number={1},
  pages={91--102},
  year={2014},
  publisher={American Physiological Society Bethesda, MD}
}

@article{grootswagers2017decoding,
  title={Decoding dynamic brain patterns from evoked responses: A tutorial on multivariate pattern analysis applied to time series neuroimaging data},
  author={Grootswagers, Tijl and Wardle, Susan G and Carlson, Thomas A},
  journal={Journal of cognitive neuroscience},
  volume={29},
  number={4},
  pages={677--697},
  year={2017},
  publisher={MIT Press One Rogers Street, Cambridge, MA 02142-1209, USA journals-info~…}
}

@book{bishop2006pattern,
  title={Pattern recognition and machine learning},
  author={Bishop, Christopher M and Nasrabadi, Nasser M},
  volume={4},
  number={4},
  year={2006},
  publisher={Springer}
}

@article{bae2018dissociable,
  title={Dissociable decoding of spatial attention and working memory from EEG oscillations and sustained potentials},
  author={Bae, Gi-Yeul and Luck, Steven J},
  journal={Journal of Neuroscience},
  volume={38},
  number={2},
  pages={409--422},
  year={2018},
  publisher={Soc Neuroscience}
}

@book{cohen2013statistical,
  title={Statistical power analysis for the behavioral sciences},
  author={Cohen, Jacob},
  year={2013},
  publisher={routledge}
}

@article{lakens2013calculating,
  title={Calculating and reporting effect sizes to facilitate cumulative science: a practical primer for t-tests and ANOVAs},
  author={Lakens, Dani{\"e}l},
  journal={Frontiers in psychology},
  volume={4},
  pages={863},
  year={2013},
  publisher={Frontiers Media SA}
}

@article{ioannidis2005most,
  title={Why most published research findings are false},
  author={Ioannidis, John PA},
  journal={PLoS medicine},
  volume={2},
  number={8},
  pages={e124},
  year={2005},
  publisher={Public Library of Science}
}

@article{kutas2011thirty,
  title={Thirty years and counting: finding meaning in the N400 component of the event-related brain potential (ERP)},
  author={Kutas, Marta and Federmeier, Kara D},
  journal={Annual review of psychology},
  volume={62},
  pages={621--647},
  year={2011},
  publisher={Annual Reviews}
}

@article{bird2000verb,
  title={Why is a verb like an inanimate object? Grammatical category and semantic category deficits},
  author={Bird, Helen and Howard, David and Franklin, Sue},
  journal={Brain and language},
  volume={72},
  number={3},
  pages={246--309},
  year={2000},
  publisher={Elsevier}
}

@article{vigliocco2011nouns,
  title={Nouns and verbs in the brain: A review of behavioural, electrophysiological, neuropsychological and imaging studies},
  author={Vigliocco, Gabriella and Vinson, David P and Druks, Judit and Barber, Horacio and Cappa, Stefano F},
  journal={Neuroscience \& Biobehavioral Reviews},
  volume={35},
  number={3},
  pages={407--426},
  year={2011},
  publisher={Elsevier}
}

@article{gouvea2010linguistic,
  title={The linguistic processes underlying the P600},
  author={Gouvea, Ana C and Phillips, Colin and Kazanina, Nina and Poeppel, David},
  journal={Language and cognitive processes},
  volume={25},
  number={2},
  pages={149--188},
  year={2010},
  publisher={Taylor \& Francis}
}

\end{document}